\documentclass[onecolumn,letterpaper]{amsart}
\pdfoutput=1
\usepackage{graphicx}

\usepackage[foot]{amsaddr}

\usepackage{amsmath}
\usepackage{amssymb}
\usepackage{color}
\usepackage[usenames,dvipsnames]{xcolor} 
\usepackage{listings}
\usepackage[font=footnotesize]{subcaption} 

\usepackage{verbatim}
\usepackage{multirow}

\usepackage{bm}

\usepackage[ruled,vlined]{algorithm2e}

\SetCommentSty{mycommfont}
\SetKwFor{ParFor}{parfor}{do}{}

\makeatletter
\newcommand\footnoteref[1]{\protected@xdef\@thefnmark{\ref{#1}}\@footnotemark}
\makeatother

\DeclareMathOperator*{\argmin}{argmin}
\DeclareMathOperator{\prox}{prox}
\DeclareMathOperator{\sgn}{sgn}
\DeclareMathOperator{\Eig}{Eig}

\renewcommand{\vec}{\bm}
\newcommand{\bbeta}{\vec{\beta}}
\newcommand{\FF}{\vec{\mathcal{F}}}
\newcommand{\EigMax}{\Eig_{\max}}

\newtheorem{definition}{Definition}[section]
\newtheorem{theorem}{Theorem}[section]
\newtheorem{remark}{Remark}[section]

\begin{document}\sloppy

\title{Feature-Adaptive Interactive Thresholding of Large 3D Volumes}

\author{Thomas Lang}
\address[T.~Lang]{FORWISS, University of Passau \& Fraunhofer Institute of Integrated Circuits, Division Development Center X-ray Technology}
\email[T.~Lang]{thomas.lang@iis.fraunhofer.de}

\author{Tomas Sauer}
\address[T.~Sauer]{Chair of Digital Image Processing; FORWISS, University of Passau \& %
Fraunhofer Institute of Integrated Circuits, Division Development Center X-ray Technology}
\email[T.~Sauer]{tomas.sauer@uni-passau.de}

%


\begin{abstract}
  Thresholding is
  the most widely used segmentation method
  in volumetric
  image processing, and its pointwise nature makes it attractive for
  the fast handling of large three-dimensional samples. However,
  global thresholds often do not properly extract components in the
  presence of artifacts, measurement noise or grayscale value
  fluctuations. 
  This paper introduces \emph{Feature-Adaptive Interactive Thresholding}
  (FAITH), a thresholding
  technique that incorporates (geometric) features, local processing
  and interactive user input to overcome these limitations.
  Given a global threshold suitable for most regions, FAITH uses
  interactively selected seed voxels to identify critical regions in
  which that threshold will be adapted locally on the basis of
  features computed from local environments around these voxels.
  The combination of domain expert knowledge and a rigorous
  mathematical model thus enables a very flexible way of local
  thresholding with intuitive user interaction. A qualitative analysis
  shows that the proposed model is able to overcome limitations
  typically occuring in plain thresholding while staying efficient
  enough to also allow the segmentation of big volumes.

\end{abstract}

\maketitle

\section{Introduction}
Volumetric segmentation is concerned with the partitioning of
three-dimensional images into disjoint regions with the goal 
to extract further information about selected components. In
particular in computed tomography or similar imaging methods,
segmentation was always a crucial part of many applications, including
clinical tomography, nondestructive inspection of parts or industrial
quality control. 

Although the underlying idea appears very simple, volumetric
segmentation is a hard problem in practice, in particular 
in modern volume processing as the increasing size of the images,
i.e., the number of volume elements, or \emph{voxels},
scales cubically with the resolution.
With ever-increasing imaging capabilities high resolution digital
volumes that do not fit into main memory anymore are generated on a
regular basis. Therefore plain computational complexity prohibits many
approaches and enforces local processing.
While pointwise thresholding does so, its limitations prevent high
quality segmentation results.

Another problem stems from the concept of generality. Ideally, one
would like to have a single segmentation 
method which can do everything for any type of image. Since this goal is unrealistic, many and foremost medical applications developed 
specialized algorithms to extract objects of a very particular type
exclusively. In this line of work, a new algorithm has to be  
designed for every object category that shall be segmented, which is
impractical for industrial purposes with their infinity of different
applications. 

In the sequel we will assume that volumes consist of voxels associated
to each 3D position having a \emph{nonnegative} value.
Our procedure will perform computations on local environments
extracted around individual voxels only. Let us formalize that.
\begin{definition}\label{def:vol}
  A \emph{volume} $V$ of dimensions $\bm{d} \in\mathbb{N}^3$ consist
  of voxels located on a regular grid having positions 
  $\bm{\alpha} \leq \bm{d}$. To each point in this grid, a voxel
  grayscale value $x_{\bm{\alpha}} \geq 0$ is 
  associated.
  The \emph{local environment} of size $K^3$ centered around the voxel
  at position $\bm{\alpha}$ is defined as the set of voxels
  \[
    \left\lbrace 
      \bm{\beta} \leq \bm{d} \mid \| \bm{\beta} - \bm{\alpha} \|_{\infty} \leq \lfloor K/2 \rfloor 
    \right\rbrace;
  \]
  for convenience, we assume that the environment size satisfies $K\in
  2\mathbb{N}+1$.
\end{definition}

The simplest and most widely used segmentation procedure is
\emph{(hard) thresholding}, a binarization based 
on a \emph{global threshold} $\theta \geq 0$ 
and defined as
\begin{equation}\label{eq:globalthresholding}
 T_{\theta}(x) = \begin{cases}1,&x \geq \theta,\\ 0,
   &\text{otherwise}.\end{cases} 
\end{equation}
For surprisingly many applications this procedure is sufficient and it
is perfectly applicable for big volumes since only information 
about the currently processed voxel is necessary for computing the result.
However, especially in industrial computed tomography artifacts and
physical effects frequently distort the grayscale values 
of certain parts of the object. In such regions, global thresholding
is typically not able to capture the desired structure and thus
produces ``holes'' there. An illustration of this effect is provided
in Figure~\ref{fig:faithskullbin} which shows such holes at the top of
the eye sockets of a skull. 
In those regions, a local binarizing threshold needs to be smaller
than the global one, but if the global threshold 
would be adapted to a lower level, other parts of the volume would
include mis-segmentations. 

The paper is organized as follows.
In Section~\ref{sec:relwork}, we briefly summarize existing approaches
to adaptive thresholding,
Section~\ref{sec:faith} introduces our model as an optimization
problem and provides an algorithm for solving this problem and we
motivate why our method is well suited for segmenting big voxel
volumes. 
In Section~\ref{sec:results} we apply our algorithm to selected
datasets and demonstrate the improvement over plain
thresholding. Finally, Section~\ref{sec:summary} summarizes our
contributions.

\section{Related Work}\label{sec:relwork}
Of course, local thresholding itself is no new idea, notable examples
include the techniques by Niblack~\cite{Niblack} 
and Sauvola~\cite{Sauvola}. Both methods choose optimal binarizing
thresholds in certain regions based on the local grayscale value distribution. However, in presence of high variance inside such regions, as it is often
the case with measurement noise in industrial computed tomography,
these methods compute a threshold which basically captures everything
regardless of any structure.
\cite{bradley} computes a local threshold by comparing voxel values to
the mean grayscale values in local regions 
and segments a voxel if it close enough to the average,
while~\cite{AMThresh} computes local thresholds in  
overlapping regions using Rosin's method, interpolates a high
resolution threshold map and performs the pointwise comparison. 
On the other hand, \cite{BoneEdgeThresh} and \cite{RATS} perform
local thresholding by combining a global 
threshold with edge detection and morphological image processing to
change the thresholding behavior locally.
\cite{CDWTThresh} computes a two-dimensional Dual Tree Complex Wavelet
Transform for each volume slice and chooses local thresholds depending
on the estimated noise level in order to reduce measurement noise.
The work in~\cite{MRAPartThresh} combines iterative image partitioning
based on thresholds with local threshold estimation. Here, after one
step of partitioning, a new local threshold is chosen from the
grayscale value distribution 
for each partition based on which a new finer partitioning is found.
An interesting yet computationally expensive approach was presented
in~\cite{VarThresh}. There, the authors 
consider the thresholding process as a pixelwise functional which,
after solving the according variational minimization
problem, segments the image.

These approaches choose a threshold for each individual voxel. On the
other hand, in practice a domain expert typically knows 
a good global threshold which is sufficient in most regions. Only few
regions require a modified threshold in order 
to detect the structures as demonstrated in~\cite{adaptiveTeeth}.
The idea of adapting a global threshold locally based on features was
implemented in~\cite{FourierFeatureThresholding}, where the authors
aim for a local adaption based on planar structure detection. However, their approach requires
two additional hyperparameters and an according tuning procedure while
only planar features influence 
the result and other structures cannot be detected.

In order to overcome these limitations, our method focusses on
allowing adaption by multiple features that can be chosen at will and
problem-dependent and works without
the need for feature-dependent hyperparameter tuning.

\section{Local Thresholding as an Optimization
  Problem}\label{sec:faith}
Our method is based on and trained from user input that is provided
interactively in such a way that several voxels are marked to identify
critical regions where the threshold needs be adapted.
Let $V$ be a volume of dimensions $\bm{d}$ as in Definition~\ref{def:vol}.
Let $S\subseteq V$ be a set of \emph{seed voxels},  
selected at $M := \#S \ll \# V$ critical positions.
Around these voxels we extract local environments $U_j$,
$j=1,\ldots,M$, of an application-dependent 
size $K\times K\times K$ where $K>1$ is odd. The size of the
environments determines the features and depends on the image, in
particular its resolution. By $F$ we denote the \emph{feature
  function} which maps local environments onto a $d$-dimensional
feature vector. Examples of such features can be found in \cite{langphd}.
In our application, $F$ is chosen depending on the dataset and
includes structural information derived from the local distribution of
grayscale values as well as geometric information
encoding the similarity of local environments to primitives including
lines and planes. It should be emphasized that \emph{any} set of local
features can be integrated in our approach.
Finally, a global threshold $\theta_g \geq 0$ needs be provided which
should be chosen such that it selects most of the object that shall be
segmented -- in plain words, it should be reasonable.

Within a local environment, we shall consider a local threshold.
\begin{definition}
  For $\theta_g \geq 0$ and  weights $\bbeta\in\mathbb{R}^d$, the
  (weighted) \emph{local 
    threshold} for a region $U$ is defined as
  \begin{equation}\label{eq:DweightedThresh}
    \theta(U) = \theta_g + \bbeta^T F(U).    
  \end{equation}
\end{definition}
\begin{remark}\label{rem1}
 With mask-based processing, extracting a full voxel environment
 around each individual voxel becomes impossible at the borders of the
 volume. Classical workarounds include periodic extension, constant
 padding, repeating or 
 mirroring the border voxel values or simply not processing them at all.
 In our application, we decided on the latter one and set the voxel
 values at the according positions explicitly zero.
\end{remark}
%
We want to choose $\bbeta$ such that the generated local thresholds
are as close to optimal local 
thresholds in the critical regions. To that end, we first compute the
features $F$ and estimate a local optimal binarizing threshold $\theta^*$ in
the regions $U_j$, $j=1,\ldots,M$.
In our concrete application, we specifically chose \emph{Minimum Cross
  Entropy Thresholding}~\cite{MCEThresholding} for $\theta^*$,
which basically computes the optimal threshold according to the cross
entropy loss for binary classification using a histogram-based procedure.
Putting everything together, we want our local thresholds to be close
to the optimal ones given the parameters. To that end, define
\begin{align*}
 \vec{\mathcal{F}} = \begin{bmatrix} F(U_1)^T \\ \vdots \\ F(U_M)^T \end{bmatrix}
 \quad \text{and} \quad
 \vec{\Theta} = \begin{bmatrix} \theta^*(U_1) - \theta_g \\ \vdots \\ \theta^*(U_M) - \theta_g \end{bmatrix},
\end{align*}
and compute the optimal feature weights $\bbeta$ 
as the solution of the minimization problem 
\[
   \argmin_{\bbeta} \frac{1}{2}\|\vec{\mathcal{F}} \bbeta - \vec{\Theta}\|_2^2.
\]
 To make the model more robust, we introduce elastic net
 regularization~\cite{ElasticNet} to this least squares problem,
 see~\eqref{eq:faithproblem}. This is known to reduce the risk of
 overfitting and only important features may adapt the threshold. 
 A trade-off between both effects is achieved by tuning its
 hyperparameters $\lambda > 0, \mu\in (0,1)$. 
 Finally, we observe that so far we would deal with an unconstrained
 optimization problem. However, if the 
 feature weights evolve in a way such that the produced local
 thresholds are always negative, we would segment
 every voxel regardless of our initial goal. To avoid this, we
 constrain the weights such that
 each local threshold is valid on the training regions. We call a
 threshold valid if it lies within the
 admissible grayscale interval $[0,W]$ where $W$ is the maximal
 representable voxel value depending on the data type, thus preventing
 thresholding with unreasonable values.
 In matrix form, define
 \begin{align*}
 \vec{C} = \begin{bmatrix} -\vec{\mathcal{F}}\\ \phantom{+}\vec{\mathcal{F}} \end{bmatrix}
 \quad \text{and} \quad
 \vec{b} = \begin{bmatrix} \;\;\phantom{(W-}\theta_g\bm{1} \\ (W-\theta_g)\bm{1} \ \end{bmatrix}
\end{align*}
which yields a closed convex polytope\\$\mathcal{C} = \{
\vec{x}\in\mathbb{R}^d \mid \vec{Cx} \leq \vec{b} \}$.

Our final optimization problem for finding the best feature weights is given as
\begin{equation}\label{eq:faithproblem}
  \begin{split}
  \argmin_{\bbeta} \quad& \frac{1}{2}\|\vec{\mathcal{F}\bbeta}-\vec{\Theta}\|_2^2 
     + \lambda\left( \frac{1-\mu}{2}\|\bbeta\|_2^2 + \mu \|\bbeta\|_1 \right)      \\
     s.t. \quad& \vec{C\bbeta} \leq \vec{b}.
 \end{split}
\end{equation}

\subsection{Solving the Problem}
We observe that the optimization problem \eqref{eq:faithproblem} can
be decomposed into
\[
  \argmin_{\bbeta} f(\bbeta) + g(\bbeta) +
  \iota_{\mathcal{C}}(\bbeta),
\]
where
\begin{align*}
 f(\bbeta) &= \frac{1}{2} \|\vec{\mathcal{F}}\bbeta -
             \vec{\Theta}\|_2^2 + \frac{\lambda(1-\mu)}{2}
             \|\bbeta\|_2^2, \\ 
 g(\bbeta) &= \lambda \mu \|\bbeta\|_1, \\ 
 \iota_{\mathcal{C}}(\bbeta) &= \begin{cases}0,
   &\bbeta\in\mathcal{C},\\+\infty, &\text{otherwise}.\end{cases} 
\end{align*} 
In this decomposition, the function $f$ is differentiable and convex
while $g$ and $\iota_{\mathcal{C}}$ are convex. It is easy to see that
all functions are proper and lower-semicontinuous as well.
Therefore, the problem can be solved using the \emph{proximal
  gradients} method, see~\cite{ProximalAlgorithms}.
To apply them on~\eqref{eq:faithproblem}, we need explicit
formulations of proximal operators which can then be interpreted as
gradient steps for convex but nonsmooth functions. In the following, we
will derive a proximal step for each of the three convex functions
and combine them into a solver for~\eqref{eq:faithproblem}.

\subsubsection{The Individual Descent Steps}
Since the convex function $f$ is differentiable, its gradient is given
explicitly by
\begin{equation}
  \label{eq:fGradient}
  \nabla f(\vec{x}) 
  = \vec{\mathcal{F}}^T \left( \vec{\mathcal{F}} \vec{x} -
    \vec{\Theta} \right) + \lambda(1-\mu)\vec{x},  
\end{equation}
which is  linear and thus Lipschitz continuous in $x$.

Concerning the nonsmooth parts of our objective function, we consider
their proximal operators, which, according to
\cite{ProximalAlgorithms} are defined as 
\[
  \prox_h(\vec{x}) = \argmin_{\vec{z}} \frac{1}{2}\|\vec{x} -
  \vec{z}\|_2^2 + h(\vec{z})
\]
for a proper lower-semicontinuous function $h$.
For our particular function $g$, it is known that the according
proximal operator is a scaled \emph{soft-thresholding} function,
cf.~\cite{SoftThresholdProximal}.
\begin{definition}
 For $\theta \geq 0$, define \emph{soft-thresholding} with respect to
 $\theta$ as
 \begin{align*}
  S_{\theta} \colon \mathbb{R}&\to\mathbb{R}, \\
  x &\mapsto \sgn(x)\,\max\{ 0, |x| - \theta \}.
 \end{align*}
\end{definition}
For the indicator function $\iota_{\mathcal{C}}$, on the other hand,
the associated proximal operator is given by the projection 
onto the polytope $\mathcal{C}$ since
\begin{align*}
 \prox_{\iota_{\mathcal{C}}}(\vec{x}) 
   &= \argmin_{\vec{z}} \frac{1}{2} \|\vec{x}-\vec{z}\|_2^2 +
     \iota_{\mathcal{C}}(\vec{z})  \\ 
   &= \argmin_{\vec{z}\in\mathcal{C}} \frac{1}{2} \|\vec{x}-\vec{z}\|_2^2
   = \mathcal{P_C}(\vec{x}).
\end{align*}
In lack of a closed-form solution for the projection onto a convex
polytope in general, we use  Hildreth's method for
inequalities~\cite[p. 283]{HildrethPolytope} to compute the projection
iteratively.

\subsubsection{Choosing the step size}
The final ingredient to iteratively train the FAITH model is the step
size whose choice it is a well-discussed topic in literature. This
includes constant step size, finding the best one in each 
iteration by a line search~\cite{BoydConvexOptimization} 
or accelerations including momentum methods~\cite{MomentumMethod}.
For simplicity, we choose a constant step size $\delta = L^{-1}$ where
$L$ denotes a Lipschitz constant of the gradient of $f$, but we
emphasize that our method can also be enhanced with the techniques
mentioned above.

By \eqref{eq:fGradient}, we have that
\begin{align*}
 \|\nabla f(\vec{x}) - \nabla f(\vec{y})\| \leq \| \FF^T \FF +
  \lambda(1-\mu)\vec{I} \|_2 \|\vec{x}-\vec{y}\|.
\end{align*}
Since adding a nonnegative multiple of the identity matrix to the
positive semidefinite $\FF^T \FF$ only shifts the eigenvalues, we note
that $\| \FF^T \FF + \lambda(1 - \mu)\vec{I} \|_2 = \| \FF^T \FF \|_2
+ \lambda (1-\mu)$ and consequently
\begin{align*}
  \| \FF^T \FF + \lambda(1 &- \mu)\vec{I} \|_2 \|\vec{x}-\vec{y}\| \\
  &= \left( \|\FF^T \FF\|_2 + \lambda(1-\mu)\|\vec{I}\|_2 \right) \|\vec{x}-\vec{y}\| \\
  &= \underbrace{\left(\EigMax(\FF^T \FF) + \lambda(1-\mu)\right)}_{=: L} \|\vec{x}-\vec{y}\|,
\end{align*}
where $\EigMax$ denotes the maximal eigenvalue of its argument matrix.
Note that $L > 0$ since $\FF^T \FF$ is positive semi-definite and
$\lambda > 0$ and $0 < \mu < 1$ holds. 

\subsubsection{Putting it Together}
The basic proximal gradient method employs \emph{forward-backward
  splitting}~\cite{SoftThresholdProximal}
which performs a gradient descent step for the differentiable part and
a proximal step for the nonsmooth part. 
Motivated this idea, one may be tempted to pursue a
\emph{forward-backward-backward} iteration of the form 
\begin{align*}
 \begin{cases}
  \bbeta^{(0)} \in \mathbb{R}^d, &\, \\
  \bbeta^{(k+1)} 
    = \mathcal{P_C}
        \left(
          S_{\delta\lambda\mu}
           \left(
             \bbeta^{(k)} - \delta \nabla f \left( \bbeta^{(k)} \right)
           \right) 
        \right). &\,
 \end{cases}
\end{align*}
While this iteration converges linearly to a unique fixed point, the
solution obtained this way is not necessarily
the optimal solution of Problem~\eqref{eq:faithproblem}. This can only be
guaranteed provided that
\begin{equation}
  \label{eq:proxRelation}
  \prox_{\delta\iota_{\mathcal{C}}+\delta g} =
 \prox_{\delta\iota_{\mathcal{C}}} \circ \prox_{\delta g} 
\end{equation}
is fulfilled, in which case the following equivalences hold:
\begin{align*}
  0 &\in \nabla f(\vec{x}^*) + \partial g(\vec{x}^*) + \partial
      \iota_{\mathcal{C}}(\vec{x}^*) \\ 
  \Leftrightarrow\; \vec{x}^* &=
                                \prox_{\delta\iota_{\mathcal{C}}+\delta
                                g}\left(\vec{x}^*-\delta\nabla
                                f(\vec{x}^*)\right)\\
  \Leftrightarrow\; \vec{x}^* &= \left(\prox_{\delta
                                \iota_{\mathcal{C}}}\circ\prox_{\delta
                                g}\right)\left(\vec{x}^*-\delta\nabla
                                f(\vec{x}^*)\right) \\
  \Leftrightarrow\; \vec{x}^* &=
                                \mathcal{P}_{\mathcal{C}}\left(S_{\delta\lambda\mu}
                                \left(\vec{x}^*-\delta\nabla 
                                f(\vec{x}^*)\right)\right) 
\end{align*}
Then, the naive fixpoint iteration converges and the fixpoint
is an optimal solution to our problem. 
However,~\eqref{eq:proxRelation} is not always fulfilled. For a counterexample,
consider $\mathcal{C} = \{ (x,y)^T \in\mathbb{R}^2 \mid y \leq -2x \}$, $\vec{\mathcal{F}}=\vec{I}$,
$\vec{\Theta}=(1,3)^T$, $\lambda=1$ and $\mu=1/2$. Let $\vec{p}=(1,3)^T$.
Then, $\delta=L^{-1}=2/3$ and we observe that
\begin{align*}
 \prox_{\iota_{\mathcal{C}}+g}(\vec{p}) 
  = \begin{bmatrix}-0.4\\0.8\end{bmatrix}
  \neq \begin{bmatrix}-0.9\\1.9\end{bmatrix}
  = \prox_{\iota_{\mathcal{C}}}\left(\prox_{g}(\vec{p})\right).
\end{align*}
To avoid this problem, we incorporate a nested optimization scheme as 
introduced in~\cite{MinimizeThreeFunctions}, and choose the following
parameters:
\begin{itemize}
 \item $(\gamma_n)_{n\in\mathbb{N}} = \delta = L^{-1}$ with $L$ as above
 \item $(\lambda_n)_{n\in\mathbb{N}} = 1$
 \item $\tau = 1$
\end{itemize}
The procedure is summarized in Algorithm~\ref{alg:nestediteration}.
We say that the algorithm has converged if the weights $\bbeta$ become
stationary, i.e., if $\|\bbeta^{(k+1)}-\bbeta^{(k)}\|_2 < \epsilon$
for a small positive number $\epsilon$.
%
\begin{algorithm}
 \caption{Iterative solver for Problem~\eqref{eq:faithproblem}.}
 \label{alg:nestediteration}
 \DontPrintSemicolon
 \SetKwInOut{Input}{Input}
 \SetKwInOut{Output}{Output}
 \SetKwFunction{FAITHTraining}{FAITH\_Training}
 \SetKwProg{Fn}{Function}{}{}
 \Fn{\FAITHTraining{$\FF$, $\theta_g$, $\lambda$, $\mu$}}{
  \Input{Feature matrix $\FF\in\mathbb{R}^{M\times d}$\\from seed voxel selection}
  \Input{Global threshold $\theta_g\geq 0$}
  \Input{Regularization parameters $\lambda > 0, \mu\in (0,1)$}
  \Output{Feature weight vector $\bbeta\in\mathbb{R}^d$}
  \BlankLine
  $\vec{C} \gets \begin{bmatrix}-\FF\\\phantom{+}\FF\end{bmatrix}$\;
  $\vec{b} \gets \begin{bmatrix} \;\;\phantom{(W-}\theta_g\bm{1} \\ (W-\theta_g)\bm{1} \ \end{bmatrix}$\;
  $\delta \gets \left(\EigMax(\FF^T \FF)+\lambda(1-\mu)\right)^{-1}$\;
  $\bbeta \gets \vec{0}$\;
  \BlankLine
  \While{$\bbeta$ not converged}{
   $\vec{x} \gets \bbeta - \delta \nabla f(\bbeta)$\;
   $\vec{z} \gets 2 S_{\delta\lambda\mu}(\vec{x}) - \vec{x}$\;
   \BlankLine
   \While{$\vec{z}$ not converged}{
    $\hat{\vec{z}} \gets \mathcal{P_C}\left(\frac{1}{2}(\vec{z}+\vec{x})\right)$\;
    $\vec{z}\gets \vec{z} + S_{\delta\lambda\mu}(2\hat{\vec{z}}-\vec{z}) - \hat{\vec{z}}$\;
   }
   \BlankLine
   $\bbeta \gets \hat{\vec{z}}$ \tcp*{The last value of $\hat{\vec{z}}$}
  }
  \Return{$\bbeta$}\;
 }
\end{algorithm}

Using the techniques from~\cite[Proposition 4.2]{MinimizeThreeFunctions}
and taking that the polytope
$\mathcal{C}$ is always nonempty as it contains at least 
the feasible point $\vec{0}$, the nested iteration can be shown in a
straightforward way to converge to the optimal solution.
\begin{theorem}
  The sequence $\bbeta^{(k)}$ generated by
  Algorithm~\ref{alg:nestediteration} converges to 
  a solution of~\eqref{eq:faithproblem}.
\end{theorem}

\subsection{Hyperparameter Tuning}
Our thresholding model contains two hyperparameters, i.e., parameters
not optimized by the solver, namely, the regularization parameters
$\lambda > 0$ and $\mu \in (0,1)$. In order to tweak them we use a
\emph{Grid Search} approach.

The hyperparameter $\mu$ controls the trade-off between sparsity and
smoothness. The closer $\mu$ approaches $1$, the more focus is laid on
sparsity while denser solutions are obtained as $\mu$ becomes zero. In
the practical application we use the discrete parameter set $P_\mu =
\{0.25,0.33,0.42,0.50,0.58,0.67,0.75\}$ and omit $\mu=1$ in
order to avoid singularities during the computation of the step size
$\delta$.

Regarding $\lambda$ we apply the approach presented
in~\cite{ElasticNetLambdaMax} where the maximum meaningful value is
given by $\lambda_{\max}(\mu) = \mu^{-1}
\|\FF^T\vec{\Theta}\|_2$. Then, a 
regularization path is constructed in log-space resulting in the parameter set
$P_\lambda = \{\epsilon \lambda_{\max}(\mu) 10^{\frac{k}{k_{\max}-1}
  \log_{10}(\epsilon^{-1})}\}_{k=0}^{k_{\max}-1}$ 
containing $k_{\max}$ values. Experimentally, we used $k_{\max}=16$
and a small $\epsilon > 0$. 

The overall hyperparameter grid is given as $P = P_\lambda \times P_\mu$.
In our application, we used $K$-fold cross validation to select the
best parameters out of this fixed set. 

\subsection{Segmenting Big Volumes}
Based on the training procedure, we can now formulate a segmentation
procedure. To that end, we consider an input volume of dimensions
$\bm{d}\in\mathbb{N}^3$ with voxels at positions
$\bm{\alpha}\leq\bm{d}$ and an output volume of identical dimensions
with voxel values $\widetilde{x}_{\vec{\alpha}}$.
The method is outlined in Algorithm~\ref{alg:faithsegmentation}.
We assume that a user, ideally domain expert, interactively selected $M$ seed voxels in
critical regions $U_1,\dots,U_M$ and furthermore provided a global
threshold $\theta_g$ that binarizes most of the components under 
investigation well.
Depending on the actual application, the user may also provide a selection 
of (geometric) features to use, but they could also be determined
automatically, cf.~\cite{langphd}.
In the first step, we extract local environments around the selected
seed voxels and compute the feature vectors on these regions to build
the feature matrix $\vec{\mathcal{F}}$. 
Then, a grid search determines the best pair of hyperparameters which
is used for training our FAITH model and obtaining feature weights
$\bbeta$.

Finally, for the eventual segmentation we iterate (possibly in
parallel) over the input volume. For each 
voxel we extract a local voxel environment around it and compute the
same set of features with proper handling of boundary voxels,
cf. Remark~\ref{rem1}.
The result is combined with the trained feature weights and the global
threshold in order to compute a localized
threshold $\theta$. The value of the voxel in the target volume at the same position as the input volume
is given as the result of binarizing thresholding with that local threshold.
\begin{algorithm}
 \caption{Segmentation using FAITH}
 \label{alg:faithsegmentation}
 \DontPrintSemicolon
 \SetKwInOut{Input}{Input}
 \SetKwInOut{Output}{Output}
 \SetKwFunction{FAITHSegment}{FAITH\_Segmentation}
 \SetKwProg{Fn}{Function}{}{}
 \Fn{\FAITHSegment{$V, S, K, F, \theta_g$}}{
  \Input{Input volume $V$}
  \Input{Seed voxel collection $S\subseteq V$, $\#S=M$}
  \Input{Environment size $K\in 2\mathbb{N}+1$}
  \Input{Feature computation function $F$}
  \Input{Global threshold $\theta_g\geq 0$}
  \Output{Output volume $\widetilde{V}$}
  \BlankLine
  $\{U_j\}_{j=1}^M \gets $ extract $K\times K\times K$ environment\\\qquad\qquad\quad{}\,around seed voxels $S$\;
  $\FF \gets \begin{bmatrix} F(U_1)\\ \vdots \\ F(U_M) \end{bmatrix}$\;
  $(\lambda^*,\mu^*) \gets $ determine by grid search\;
  $\bbeta \gets $ \FAITHTraining{$\FF, \theta_g, \lambda^*, \mu^*$}\;
  \BlankLine
  \For{$\bm{\alpha}$ in $V$}{
    $U \gets $ $K\times K\times K$ environment around $\bm{\alpha}$\;
    $\theta \gets \theta_g + \bbeta^T F(U)$\;
    $\widetilde{x}_{\bm{\alpha}} \gets T_\theta(x_{\bm{\alpha}})$\;
  }
 }
\end{algorithm}

\subsection{Runtime and Memory Analysis}
For large voxel volumes, only algorithms with linear runtime in the
number of voxels are admissible in practice. Therefore, we briefly
analyze the runtime and memory requirements of our thresholding
procedure.

Concerning the runtime, the extraction of local environments around
the seed voxels has a runtime of $\mathcal{O}(M)$ where $M$ denotes
the number of seed voxels. 
The training itself consists of a nested solver loop with $m$ iterations of
the outer loop and $k_j, j=1,\ldots,m$, the associated inner iterations. 
With $k = \sum_{j=1}^m k_j$, a single FAITH training has
a runtime complexity of $\mathcal{O}(k)$. Each iteration depends on the number
of seed voxels $M$ in some polynomial way, however since $M$ is
typically very small and, more important, \emph{independent} of the number
of voxels in the volume, the according factor will be so as well.
Aggregating the runtimes of the training involving a grid search over
$\#P$ parameters using $t$ threads and 
the last training with the found hyperparameters, we obtain
$\mathcal{O}(k(1+\frac{\#P}{t}))$. 
The final step involves a single pass through the voxel volume which
is clearly linear in the number $N$ of voxels.

Overall, the runtime complexity of
Algorithm~\ref{alg:faithsegmentation} can thus be estimated as 
\[
  \mathcal{O}\left( M + k\left(1+\frac{\#P}{t}\right) + N \right).
\]
Already for practical reasons, the number of seed voxels $M$ is
typically very small (since they have to be marked manually) and
independent of $N$, thus asymptotically yielding an effort which is
linear with respect to the number of voxels $N$.

Considering memory requirements, we notice that during training we need to keep the feature matrix, the polytope
constraints and of course the solution vector in memory. The feature
matrix $\vec{\mathcal{F}}$ is of size $M\times d$, where $d$ is the
dimensionality of the feature vector.
Likewise, the polytope constraints consist of a matrix $\vec{C}$ of
size $2M\times d$ and a  
vector $\vec{b}$ of size $2M$. The solution vector $\bbeta$ has
dimensionality $d$. 
Thus, the overall space complexity for training scales with $Md$.
The classification phase only relies on local environments of fixed size and the trained solution vector,
thus having constant memory complexity.
Fortunately, the feature vector dimensionality is constant after feature selection and thus the memory
requirements scale linearly with the number of seed voxels which is assumed independent of the number of
voxels, making our segmentation procedure especially well suited for
handling large three-dimensional datasets.

\section{Results}\label{sec:results}
\begin{figure*}
 \centering
 \begin{subfigure}[b]{.32\textwidth}
  \centering
  \includegraphics[width=\linewidth]{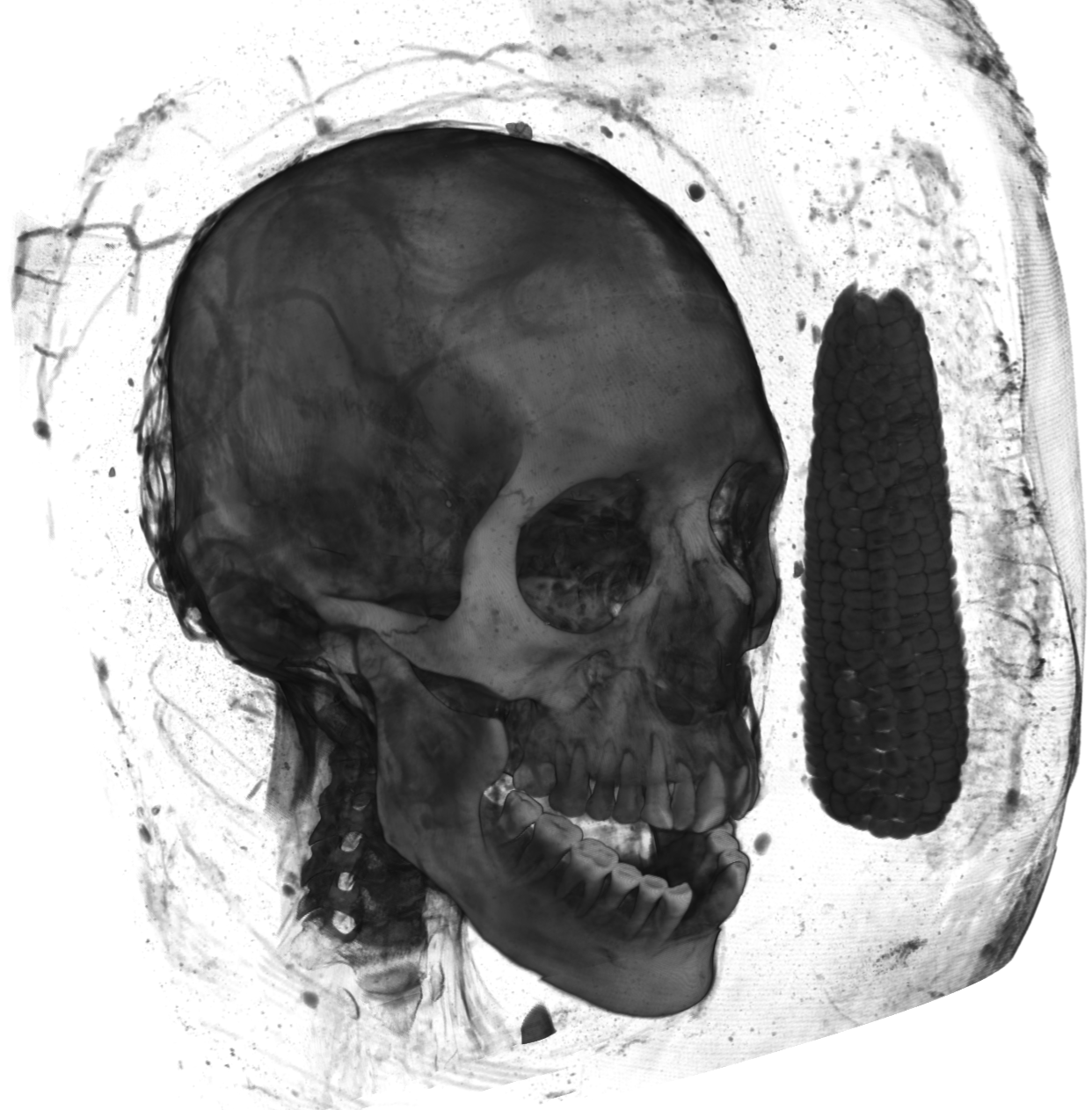}
  \caption{Mummy head}
  \label{fig:mummyorig}
 \end{subfigure}
 \begin{subfigure}[b]{.21\textwidth}
  \centering
  \includegraphics[width=\linewidth]{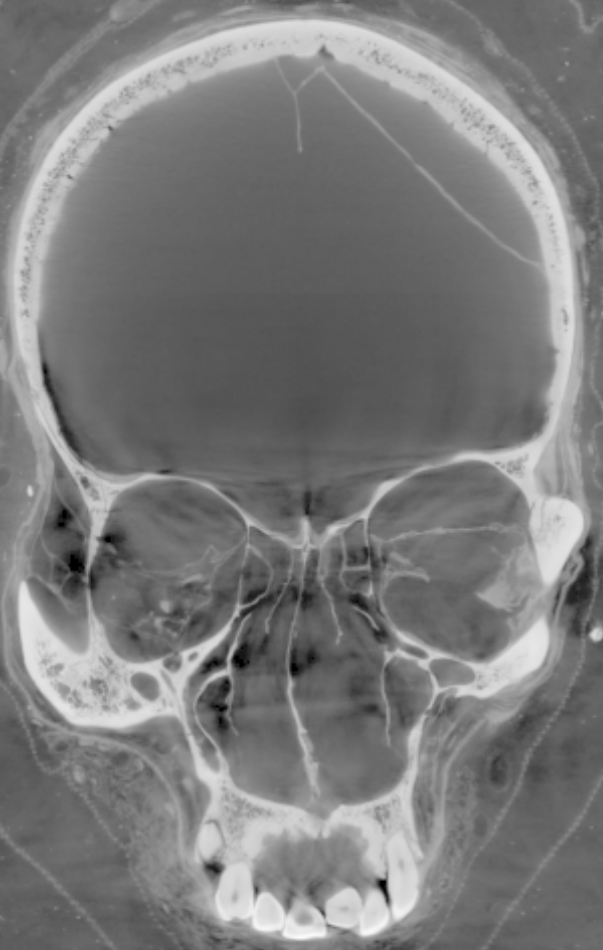}
  \caption{Original data}
  \label{fig:faithskullorig}
 \end{subfigure}
 \begin{subfigure}[b]{.21\textwidth}
  \centering
  \includegraphics[width=\linewidth]{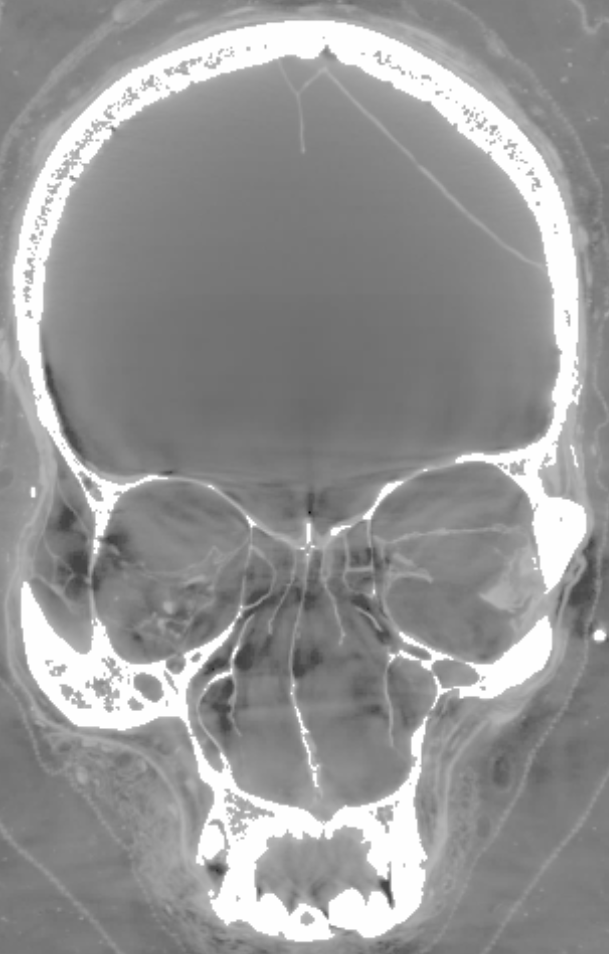}
  \caption{Binarization}
  \label{fig:faithskullbin}
 \end{subfigure}
 \begin{subfigure}[b]{.21\textwidth}
  \centering
  \includegraphics[width=\linewidth]{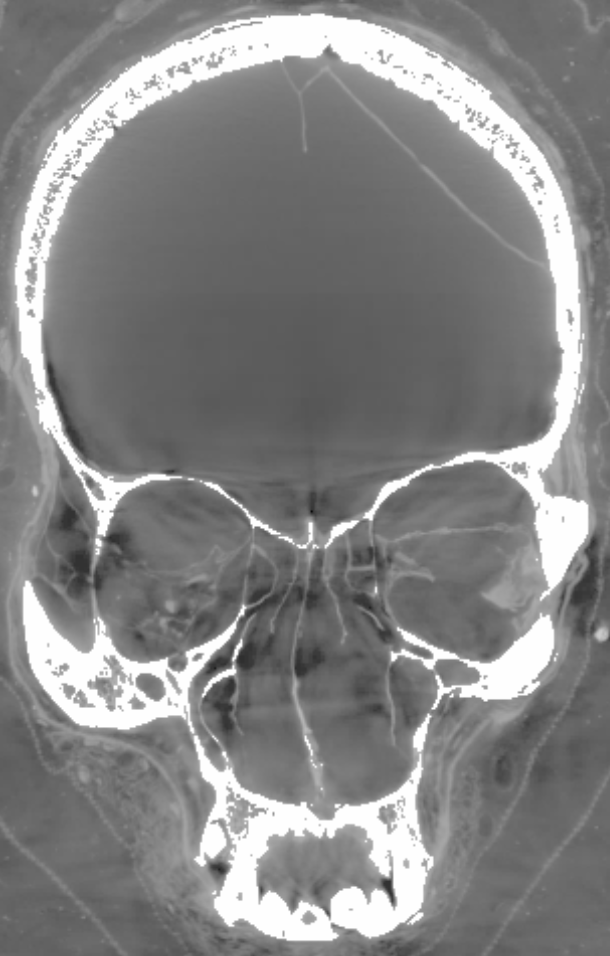}
  \caption{FAITH}
  \label{fig:faithskull}
 \end{subfigure}
 \caption{Binarization applied to the mummy skull data in~(\subref{fig:faithskullorig}) produces holes 
         at the top of the eye sockets~(\subref{fig:faithskullbin}).
         FAITH is able to detect bone in these critical sections while the overall segmentation remains 
         good~(\subref{fig:faithskull}).
         Results are overlaid in white over the original data.}
 \label{fig:faithapplied}
\end{figure*}

\begin{figure*}
 \centering
 \begin{subfigure}[b]{.22\textwidth}
  \centering
  \includegraphics[width=\linewidth]{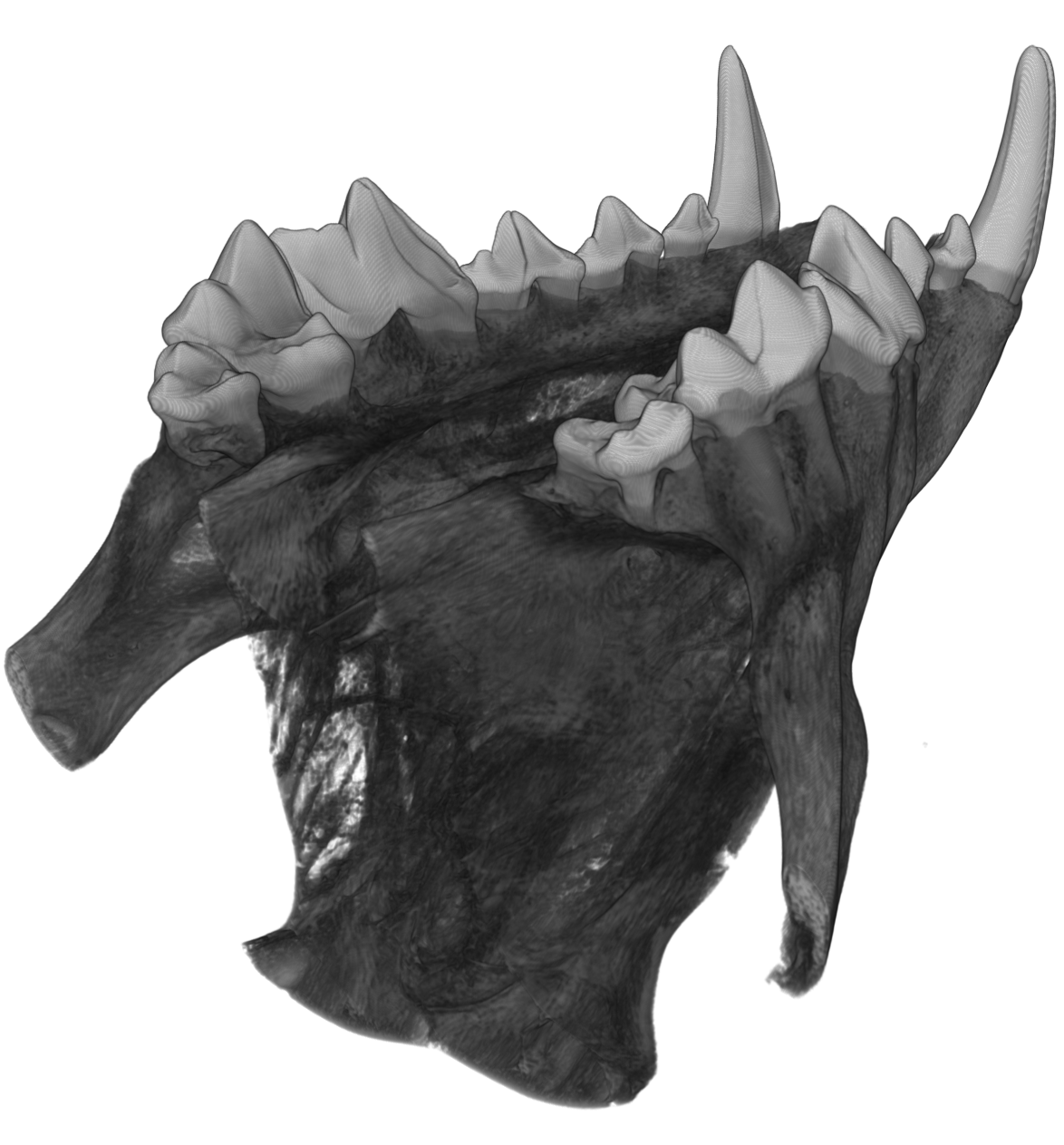}
  \caption{Wolf jaw}
  \label{fig:canislupusorig3d}
 \end{subfigure}
 \begin{subfigure}[b]{.25\textwidth}
  \centering
  \includegraphics[width=\linewidth]{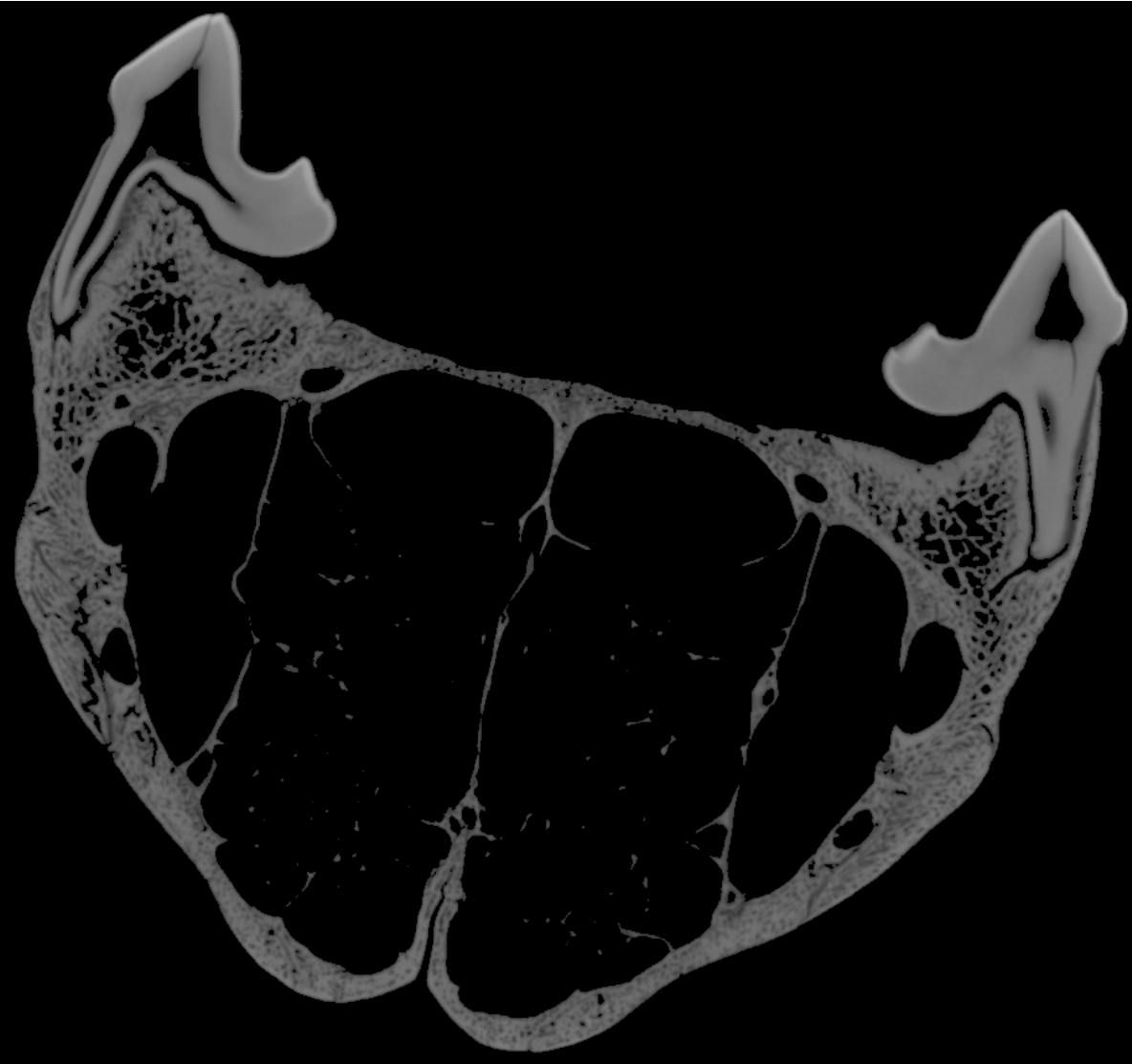}
  \caption{Original data}
  \label{fig:canislupusorig}
 \end{subfigure}
 \begin{subfigure}[b]{.25\textwidth}
  \centering
  \includegraphics[width=\linewidth]{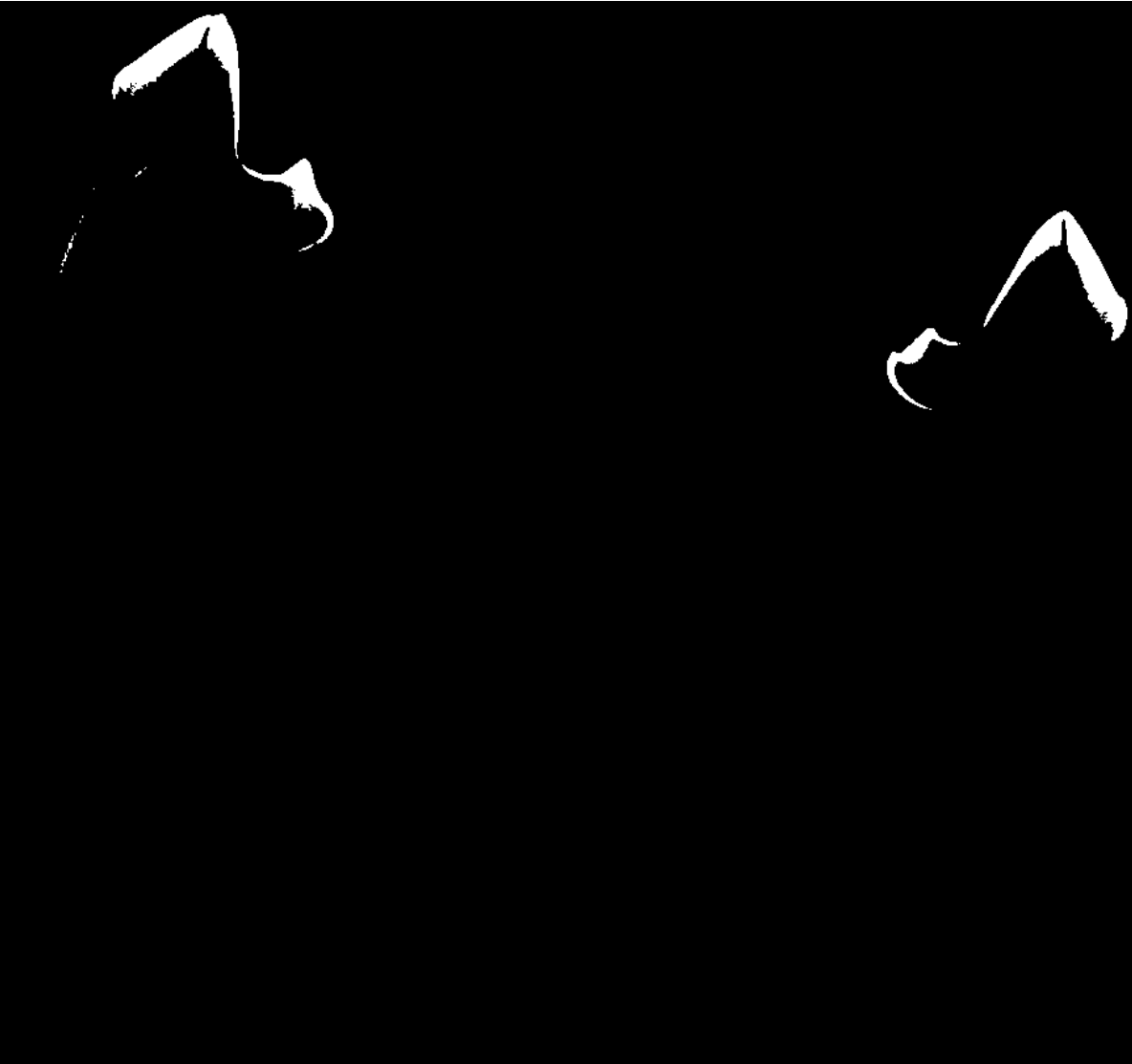}
  \caption{Binarization}
  \label{fig:canislupusbin}
 \end{subfigure}
 \begin{subfigure}[b]{.25\textwidth}
  \centering
  \includegraphics[width=\linewidth]{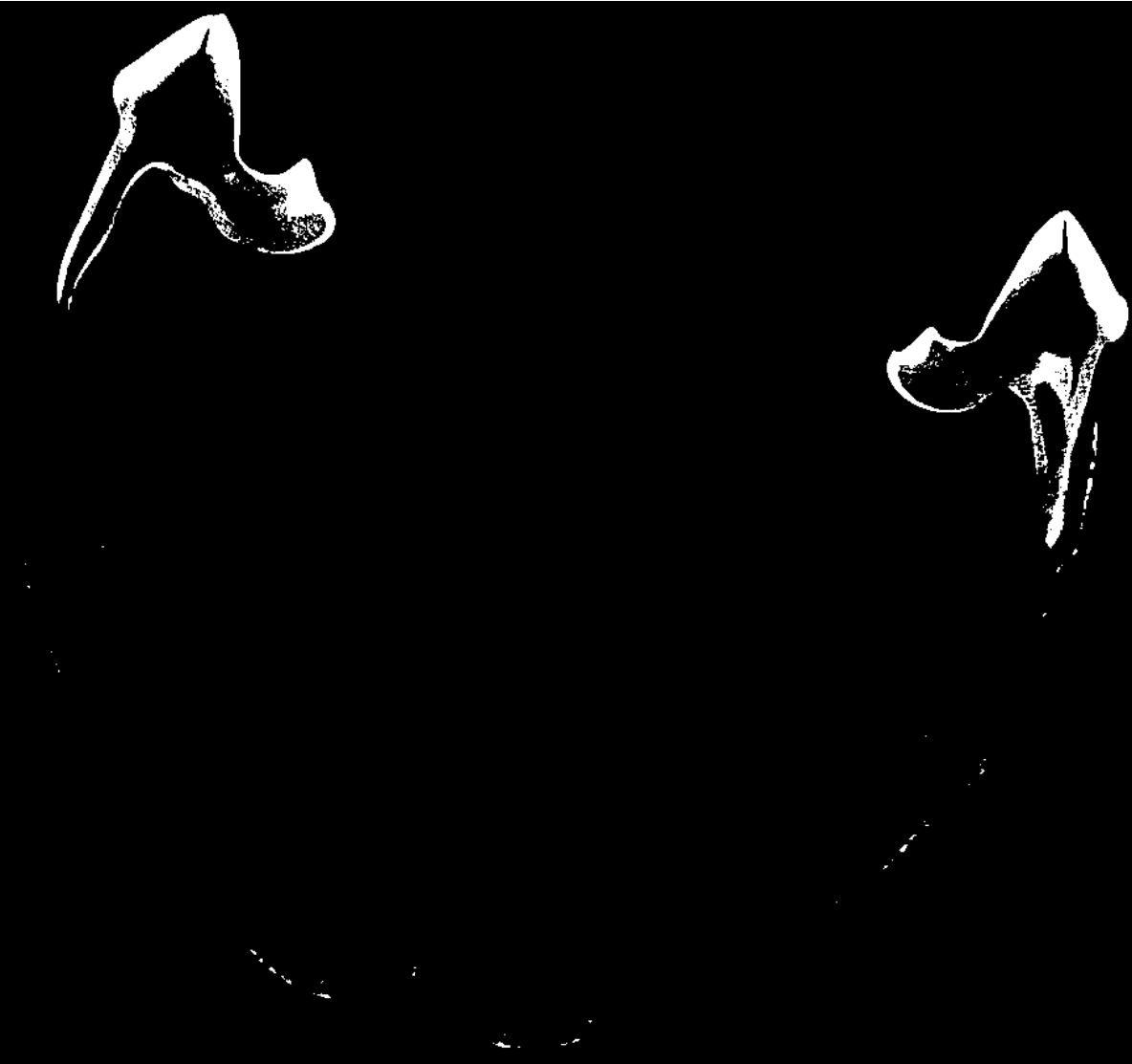}
  \caption{FAITH}
  \label{fig:canislupusfaith}
 \end{subfigure}
 \caption{Binarization applied to a scan of wolf jaw labeled \emph{Canis Lupus} in~(\subref{fig:canislupusorig}) discards most 
          of the teeth structure, see~(\subref{fig:canislupusbin}).
          Our procedure detects the teeth better while simultaneously discarding the jaw bone
          we wish to ignore~(\subref{fig:faithskull}).}
 \label{fig:canislupusfaithall}
\end{figure*}

In this section we demonstrate the capabilities of our algorithm to
detect geometric structures not captured by global thresholding by
means of two examles.
Figure~\ref{fig:mummyorig} shows a rendering of the head section of a
Peruvian mummy now located at 
the Lindenmuseum in Stuttgart, while Figure~\ref{fig:canislupusorig3d}
shows a computed tomography scan of a wolf jaw labeled "Canis Lupus". 
Both scans were generated by the Fraunhofer Development Center for
X-ray Technology in Fürth.

\begin{figure*}
 \centering
 \begin{subfigure}[b]{.25\textwidth}
  \centering
  \includegraphics[width=\linewidth]{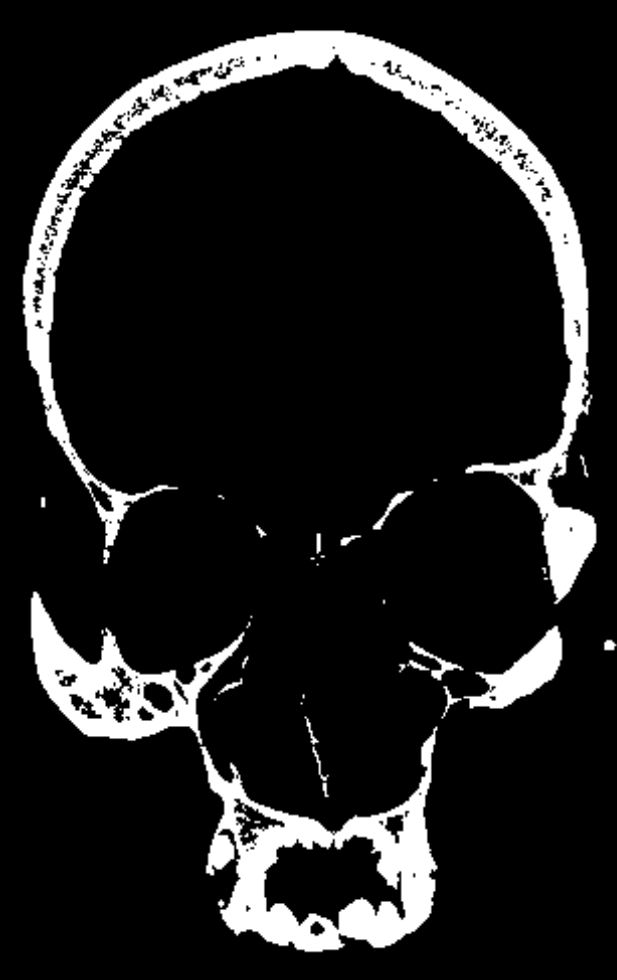}
  \caption{Binarization}
  \label{fig:faithskullplain}
 \end{subfigure}
 \begin{subfigure}[b]{.25\textwidth}
  \centering
  \includegraphics[width=\linewidth]{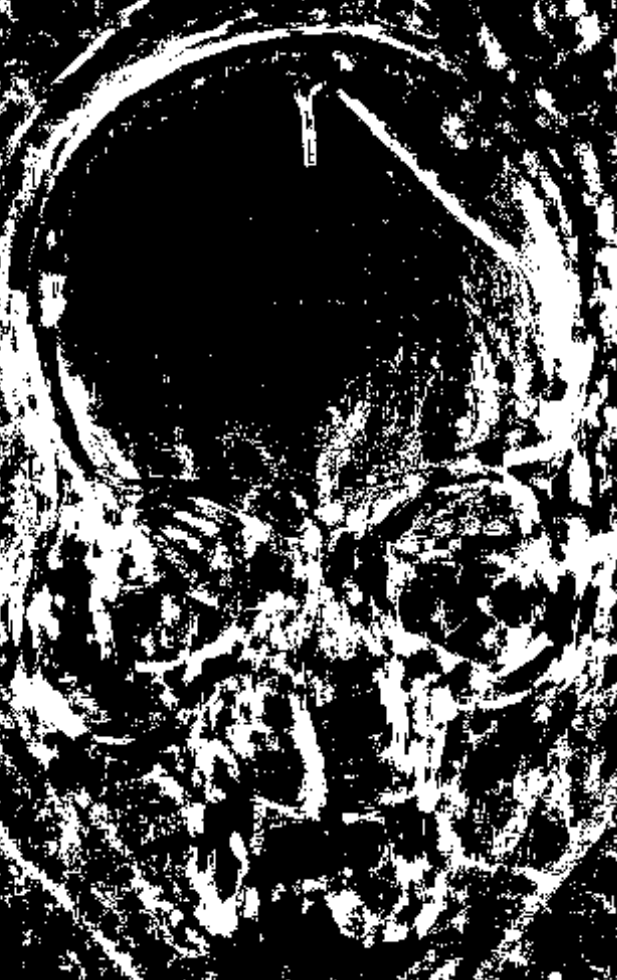}
  \caption{Least Squares only}
  \label{fig:faithskullls}
 \end{subfigure}
 \begin{subfigure}[b]{.25\textwidth}
  \centering
  \includegraphics[width=\linewidth]{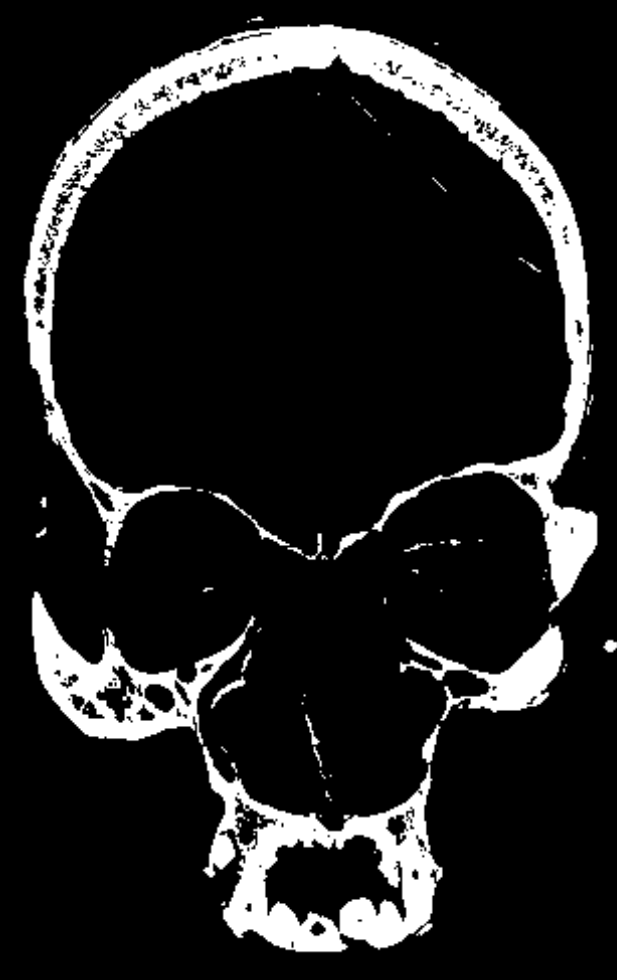}
  \caption{FAITH}
  \label{fig:faithskullfull}
 \end{subfigure}
 \caption{Applying global binarization fails to detect bone in the upper eye socket 
          regions~(\subref{fig:faithskullplain}).
          FAITH without any kind of regularization and validity constraints, i.e., using only 
          a simple least squares approximation, produces a very cluttered result~(\subref{fig:faithskullls}).
          The full procedure omits these problems and shows the better segmented skull in~(\subref{fig:faithskullfull}).
 }
 \label{fig:faithregseries}
\end{figure*}

Our first aim was to segment the skull of the mummy without capturing
the wrappings around it. Doing so,
we used the global threshold specified in Table~\ref{table:setup}. In
Figure~\ref{fig:faithskullbin} 
we see that global thresholding results in holes at the top of the eye
sockets, whereas lowering that threshold would 
result in segmenting the wrappings too. We applied FAITH
by interactively selecting 54 seeds out of 104,368,576 total voxels
in these critical regions and set the parameters as in Table~\ref{table:setup},
using the \emph{same} global threshold. The result is given in
Figure~\ref{fig:faithskull} with a clearly visible improvement in the
critical regions. At the same time, the rest of the skull which was 
detected properly by the global threshold stays unaffected. Thus, the
\emph{planar} structures of the eye socket are detected by the
features quite well and the segmentation result improved
considerably. 
As a side note we like to demonstrate the influence of our
modifications to the basic least squares approximation. 
The differences can be observed in
Figure~\ref{fig:faithregseries}. The least squares problem without
regularization is rendered ill-conditioned if a feature produces the
very same values for multiple environments, resulting in a volume
cluttered with missegmentations. Regularization and restricting to
valid thresholds improves this to the known result. 

We also applied our procedure to the wolf jaw scan. Here, we specified
the threshold in a way such that the teeth 
are kept as good as possible, while the jaw bone shall be
discarded. Unfortunately, binarization using a
global threshold also discards most of the teeth,
cf. Figure~\ref{fig:canislupusbin}.
Again, we interactively selected critical regions, set the FAITH
parameters as specified in 
Table~\ref{table:setup} and applied the segmentation method.
Here, we selected 166 out of 2,467,050,300 total voxels.
The result is depicted in Figure~\ref{fig:canislupusfaith} in which the teeth are detected better. While
individual voxels of the jaw bone remain, these can be easily removed by simple postprocessing.

The execution times of these applications are summarized in
Table~\ref{table:runtime}. The implementation is within a Fraunhofer
software framework and runs exclusively on the CPU; the measurements
were recorded on an  
Intel\textsuperscript{\textregistered}
Core\textsuperscript{\texttrademark} i7-770H processor  
with eight logical cores and an average clock rate of 4.2~GHz.
While the procedure is algorithmically efficient in the sense that
its runtime is linear with respect to the number of voxels, the cubic
increase of the number of voxels with respect to resolution makes us
propose that in future work this algorithm should be implemented
on GPUs to accelerate this even further.

\begin{table}
 \centering
 \begin{tabular}{|l||r|c|c|}
  \hline
  \textbf{Scan} & \textbf{Threshold} & \textbf{Env. size} & \textbf{\#Features} \\ \hline
  Canis Lupus   &              24415 &                  5 & \multirow{2}{*}{2}  \\ \cline{1-3}
  Mummy skull   &               1541 &                  7 &                     \\ \hline
 \end{tabular}
 \caption{Experiment setup}
 \label{table:setup}
\end{table}

\begin{table}
 \centering
 \begin{tabular}{|l||r|r|r|}
  \hline 
  \textbf{Scan} & \textbf{Size} & \textbf{Time} \\ \hline
  Mummy skull &  199~MiB & 1801.79~s  \\ \hline
  Canis Lupus & 4076~MiB & 3155.80~s  \\ \hline
 \end{tabular}
 \caption{Runtime measurements}
 \label{table:runtime}
\end{table}

\section{Conclusions}\label{sec:summary}
In this work we proposed FAITH, a novel thresholding technique which
adapts a global threshold locally on the basis of freely selectable features. 
The main benefit is that a domain expert can interactively identify
critical regions where global thresholding 
produces undesirable results. Inside these regions, features extract structural
information and a model 
is trained to modify the global threshold in a way such that the
localized threshold is optimal
inside these regions. At the same time, the global one is modified only if the trained structure
is observed, thus in regions where the global threshold suffices for segmentation,
the detection quality remains the same.
Overall, this procedure allows improving the segmentation quality over
global thresholding  
interactively and at the same time handling three-dimensional datasets
without size  
restrictions due to purely local processing.
Furthermore, the same algorithm can be used for a variety of different
domains and applications as long as we 
process volumes and the object under investigation can be segmented
using a thresholding approach. 

\medskip
\textbf{Acknowledgements}
This work was supported by the project 
"Digitalisierung, Verarbeitung und Analyse kultureller und industrieller
Objekte: Wertschöpfung aus großen Daten und Datenmengen - Big Picture" that has been promoted and funded by
the Bavarian State Government under Grant No. AZ.: 43-6623/138/2 from 2018 to 2021.

\bibliographystyle{IEEEtran}
\bibliography{bibliography}

\begin{thebibliography}{10}
\providecommand{\url}[1]{#1}
\csname url@samestyle\endcsname
\providecommand{\newblock}{\relax}
\providecommand{\bibinfo}[2]{#2}
\providecommand{\BIBentrySTDinterwordspacing}{\spaceskip=0pt\relax}
\providecommand{\BIBentryALTinterwordstretchfactor}{4}
\providecommand{\BIBentryALTinterwordspacing}{\spaceskip=\fontdimen2\font plus
\BIBentryALTinterwordstretchfactor\fontdimen3\font minus
  \fontdimen4\font\relax}
\providecommand{\BIBforeignlanguage}[2]{{%
\expandafter\ifx\csname l@#1\endcsname\relax
\typeout{** WARNING: IEEEtran.bst: No hyphenation pattern has been}%
\typeout{** loaded for the language `#1'. Using the pattern for}%
\typeout{** the default language instead.}%
\else
\language=\csname l@#1\endcsname
\fi
#2}}
\providecommand{\BIBdecl}{\relax}
\BIBdecl

\bibitem{Niblack}
W.~Niblack, \emph{An Introduction to Digital Image Processing}.\hskip 1em plus
  0.5em minus 0.4em\relax Birkeroed, Denmark, Denmark: Strandberg Publishing
  Company, 1985.

\bibitem{Sauvola}
J.~Sauvola and M.~Pietikäinen, ``Adaptive document image binarization,''
  \emph{Pattern Recognit.}, vol.~33, no.~2, pp. 225--236, 2000.

\bibitem{bradley}
D.~Bradley and G.~Roth, ``{Adaptive Thresholding using the Integral Image},''
  \emph{J. Graph. Tools}, vol.~12, pp. 13--21, 01 2007.

\bibitem{AMThresh}
J.~Lifton and T.~Liu, ``{An adaptive thresholding algorithm for porosity
  measurement of additively manufactured metal test samples via X-ray computed
  tomography},'' \emph{Addit. Manuf.}, vol.~39, p. 101899, 2021.

\bibitem{BoneEdgeThresh}
A.~J. Burghardt, G.~J. Kazakia, and S.~Majumdar, ``{A Local Adaptive
  Thresholding Strategy for High Resolution Peripheral Quantitative Computed
  Tomography of Trabecular Bone},'' \emph{Ann. Biomed. Eng.}, vol.~35, no.~10,
  pp. 1678--1686, 2007.

\bibitem{RATS}
J.~Kittler, J.~Illingworth, and J.~Föglein, ``Threshold selection based on a
  simple image statistic,'' \emph{Comput. Vis. Graph. Image Process.}, vol.~30,
  no.~2, pp. 125--147, 1985.

\bibitem{CDWTThresh}
M.~Diwakar, Sonam, and M.~Kumar, ``{CT Image Denoising Based on Complex Wavelet
  Transform Using Local Adaptive Thresholding and Bilateral Filtering},'' in
  \emph{Proceedings of the Third International Symposium on Women in Computing
  and Informatics}, ser. WCI '15, 2015, pp. 297–--302.

\bibitem{MRAPartThresh}
D.~Y. Kim and J.~W. Park, ``{Connectivity-Based Local Adaptive Thresholding for
  Carotid Artery Segmentation Using MRA Images},'' \emph{Image Vis. Comput.},
  vol.~23, no.~14, pp. 1277--–1287, 12 2005.

\bibitem{VarThresh}
F.~H. Chan, F.~Lam, and H.~Zhu, ``{Adaptive Thresholding by Variational
  Method},'' \emph{IEEE Trans. Image Process.}, vol.~7, no.~3, pp. 468--473,
  1998.

\bibitem{adaptiveTeeth}
J.~Michetti, A.~Basarab, F.~Diemer, and D.~Kouame, ``Comparison of an adaptive
  local thresholding method on {CBCT} and $\mathrm{\mu}${CT} endodontic
  images,'' \emph{Phys. Med. Biol.}, vol.~63, no.~1, p. 015020, 2017.

\bibitem{FourierFeatureThresholding}
C.-F. Westin, A.~Bhalerao, H.~Knutsson, and R.~Kikinis, ``Using {L}ocal 3{D}
  {S}tructure for {S}egmentation of {B}one from {C}omputer {T}omography
  {I}mages,'' in \emph{CVPR}, 06 1997, pp. 794--800.

\bibitem{langphd}
T.~Lang, ``{AI-Supported Interactive Segmentation of 3D Volumes},'' Ph.D.
  dissertation, University of Passau, 07 2021.

\bibitem{MCEThresholding}
C.~Li and C.~Lee, ``{Minimum cross entropy thresholding},'' \emph{Pattern
  Recognit.}, vol.~26, no.~4, pp. 617 -- 625, 1993.

\bibitem{ElasticNet}
H.~Zou and T.~Hastie, ``{Regularization and variable selection via the elastic
  net},'' \emph{J. R. Stat. Soc., Series B}, vol.~67, pp. 301--320, 2005.

\bibitem{ProximalAlgorithms}
N.~Parikh and S.~Boyd, ``{Proximal Algorithms},'' \emph{Found. Trends Optim.},
  vol.~1, no.~3, pp. 127--239, 01 2014.

\bibitem{SoftThresholdProximal}
P.~Combettes and V.~R.~Wajs, ``{Signal Recovery by Proximal Forward-Backward
  Splitting},'' \emph{Multiscale Model Simul.}, vol.~4, 01 2005.

\bibitem{HildrethPolytope}
G.~T. Herman and A.~Lent, ``{Iterative reconstruction algorithms},''
  \emph{Comput. Biol. Med.}, vol.~6, no.~4, pp. 273--294, 1976.

\bibitem{BoydConvexOptimization}
S.~Boyd and L.~Vandenberghe, \emph{{Convex Optimization}}.\hskip 1em plus 0.5em
  minus 0.4em\relax Cambridge University Press, 03 2004.

\bibitem{MomentumMethod}
D.~E. Rumelhart, G.~E. Hinton, and R.~J. Williams, \emph{{Learning
  Representations by Back-Propagating Errors}}.\hskip 1em plus 0.5em minus
  0.4em\relax Cambridge, MA, USA: MIT Press, 1988, p. 696–699.

\bibitem{MinimizeThreeFunctions}
C.~Chaux, J.-C. {P}esquet, and N.~{P}ustelnik, ``Nested {I}terative
  {A}lgorithms for {C}onvex {C}onstrained {I}mage {R}ecovery {P}roblems,''
  \emph{SIAM J. Imaging Sci.}, vol.~2, no.~2, pp. 730--762, 2009.

\bibitem{ElasticNetLambdaMax}
R.~Tibshirani, T.~Hastie, and J.~Friedman, ``{Regularized Paths for Generalized
  Linear Models Via Coordinate Descent},'' \emph{J. Stat. Softw.}, vol.~33, 02
  2010.

\end{thebibliography}

\end{document}